# Machine Learning Methods to Analyze Arabidopsis Thaliana Plant Root Growth

*Hamidreza Farhidzadeh

*Department of Mathematics and Computer Science, Amirkabir University of Technology, Tehran, Iran

**Abstract-** One of the challenging problems in biology is to classify plants based on their reaction on genetic mutation. Arabidopsis Thaliana is a plant that is so interesting, because its genetic structure has some similarities with that of human beings. Biologists classify the type of this plant to mutated and not mutated (wild) types. Phenotypic analysis of these types is a time-consuming and costly effort by individuals. In this paper, we propose a modified feature extraction step by using velocity and acceleration of root growth. In the second step, for plant classification, we employed different Support Vector Machine (SVM) kernels and two hybrid systems of neural networks. Gated Negative Correlation Learning (GNCL) and Mixture of Negatively Correlated Experts (MNCE) are two ensemble methods based on complementary feature of classical classifiers; Mixture of Expert (ME) and Negative Correlation Learning (NCL). The hybrid systems conserve of advantages and decrease the effects of disadvantages of NCL and ME. Our Experimental shows that MNCE and GNCL improve the efficiency of classical classifiers, however, some SVM kernels function has better performance than classifiers based on neural network ensemble method. Moreover, kernels consume less time to obtain a classification rate.

## INTRODUCTION

In biology, study on root growth and organisms operation is a significant issue. Understanding of effective factors on organism operation can be useful in various fields, from treatment of diseases up to increasing the fertility of plants. Studies about usage of organisms' morphological characteristics to obtain information about genetic operation of these organisms are called phenotypic analysis.

Many genetic factors effect on each morphological characteristics of an organism. For instance, the manner of gene operation can be discovered with comparison of differences between two organisms that are distinct in a specific gene. Technology advances in the field of obtaining genetic characteristics and morphological characteristics improve phenotypic analysis. Because of high amount of data and required accuracy in these studies, there is increasing demand for fast and accurate computational methods to extract phenotypic information and distinguish between different species.

Unlike the past that an extensive collection of organisms in biological studies was investigated, only a few special organisms are reviewed for such researches now. The reason of this issue is to simplify the experiments, to reduce the required time and cost of experiments. The Arabidopsis Thaliana plant is one of the instances that have attracted a lot of research interest, recently. Arabidopsis Thaliana is a small plant that grows in short period and by possessing 5 chromosomes and 150 base pairs, it has one of the smallest genome between plants. It is the first plant that its genetic sequence is determined [1]. In this plant, two specific genes are mutated and the behaviours of mutated species and wild type species are analysed.

We target to apply a new approach for feature extraction and classification. We apply velocity and acceleration of root growth and use sliding window to decrease classification error rate. We also obtain classification accuracy rate by classifiers combination. Combining classifiers is one of the methods to improve performance in classification issues, especially, in complicated cases that have few numbers of training data, feature vector with high dimension or overlapped classes [12]. In accordance to experiments results, "divide and conquer" idea improves performance of classification by dividing main problem to some easier computational problems and then by combining the results. In supervised learning classification issues, divide and conquer principle is implemented by separating data space to some sub-problems and attributing the experts to model each sub problems [7].

The rest of this article is organized as follows. Section 2 reviews the SVM methods and its kernels. Section 3 presents some related works are done on Arabidopsis Thaliana root growth and analysis of swing of tip angle. In the Section 4 the learning and combining procedure of the ME and NCL methods are investigated. In Section 5, the new feature extraction approach is explained. Section 6 presents the results of our experimental study on Arabidopsis Thaliana plant. Finally, Section 6 concludes the article.

REVIEW OF SVM AND KERNEL FUNCTIONS

The important issue in classification is to find the best decision boundary. We can suppose a two-class problem that is linearly separable. We can find out there is many decision boundaries. However, the main question is: Are all decision boundaries equally efficient?

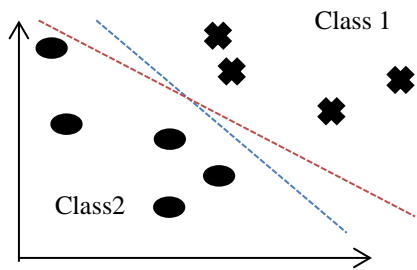

Support Vector Machines (SVMs) are a class of supervised learning algorithms first introduced by Vapnik[?? Therodosis] . Given a set of labeled training vectors (positive and negative input examples), SVMs learn a linear decision boundary to discriminate between the two classes. The result is a linear classification rule that can be used to classify new test examples. SVMs have exhibited excellent generalization performance (accuracy on test sets) in practice and have strong theoretical motivation in statistical learning theory 19.

Let $X$ with $\{x_1, x_2... x_n\} \in R^n$ be our data set and $y_i \in \{-1,+1\}$ be the class label of $x_i$. We can specify a linear classification rule $f$ by a pair $(w, b)$,

$$f(x) = w^T x + b, \qquad (1)$$

where $w \in R^n$ and $b \in R^n$, a point x is classified as positive (negative) if $f(x) > 0$ ($f(x) < 0$). Geometrically, the decision boundary is the hyper plane

$$\{x \in R^n: w^T x + b = 0\} \qquad (2)$$

where $w$ is a normal vector to the hyper plane and b is the bias. The margin $m$ of the classifier with respect to our data set members is:

$$m = Min_{x_i}(y_i(w^T x + b)) \qquad (3)$$

When classes are linearly separable, the classifier can classify the dataset and m is the distance of decision boundary and nearest training points.

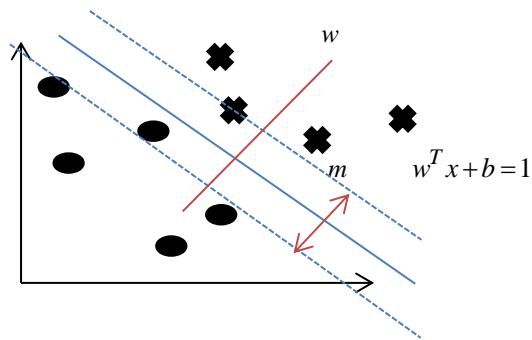

$$w^T + b = 0$$

$$w^T x + b = -1$$

One types of SVM is that appropriate decision boundary has maximum margin *m* of both classes' data as far as possible. Therefore, the SVM discovers the decision boundary that separates data and maximizes the distance to nearest training points.

However, generally, training sets are not linearly separable and we should enhance the SVM efficiency to balance a trade-off between maximizing margin and minimizing classification error rate on training set.

*Kernels in SVMs*

A key feature of any SVM optimization problem is that it is equivalent to solving a dual quadratic programming problem. For example, in the linearly separable case, the maximal margin classifier is found by solving for the optimal "weights" $\alpha_i$, I = 1,…, m, in the dual problem:

$$\text{Maximize} \quad W(\alpha) = \sum_{i=11}^{n} \alpha_i - \frac{1}{2} \sum_{i=1, j=1}^{n} \alpha_i \alpha_j y_i y_j x_i^T x_j \quad (4)$$

$$\text{Subject} \quad \alpha_i \geq 0 \;,\; \sum \alpha_i y_i = 0$$

*w* and *b* of classifier are then determined by the optimal $\alpha_i$. The dual problem not only modifies SVMs to various efficient optimization algorithms, but also, since the dual problem depends only on the inner products $x_i$ and $x_j$ allows for the introduction of kernel techniques.

The key idea is to transform $x_i$ to a higher dimensional space to make classification easier. The input space is space that $x_i$ is belong to. The feature space is a space that any given feature map $\Phi(x_i)$ from *X* to higher dimensional vector space.

$$\Phi : X \rightarrow R^n$$

So, the kernel K on X*X is as follows:

$$K(x_i, x_j) = \Phi^T(x_i) * \Phi(x_i) \quad (5)$$

We can use SVM in feature vector by replacing inner products with kernel function in dual programming.

*Gaussian and Sigmoid Kernel*

We need a simple way to test whether a function constitutes a valid kernel without having to construct the function Φ (x) explicitly. A necessary and sufficient condition for a function $K(x_i, x_j)$ to be a valid kernel (Shawe-Taylor and Cristianini, 2004) is that the Gram matrix *K*, whose elements are given by $k(x_i, x_j)$, should be positive semi definite for all possible choices of the set $\{x_n\}$. Note that a positive semi definite matrix is not the same thing as a matrix whose elements are nonnegative.

We require that the kernel $K(x_i, x_j)$ be symmetric and positive semidefinite and that it expresses the appropriate form of similarity between x and x_ according to the intended application.

One of commonly used kernel takes the form

$$K(x_i, x_j) = \exp(-||x_i - x_j||^2 / 2\sigma^2) \quad (6)$$

and is often called a 'Gaussian' kernel. Note, however, that in this context, it is not interpreted as a probability density, and hence the normalization coefficient is omitted. We can see that this is a valid kernel by expanding the square:

$$||x_i - x_j||^2 = x_i^T x_i + x_j^T x_j - 2x_i^T x_j \tag{7}$$

Note that the feature vector that corresponds to the Gaussian kernel has infinite dimensionality. The Gaussian kernel is not restricted to the use of Euclidean distance. If we use kernel substitution in (6.24) to replace $x_i^T x_j$ with a nonlinear kernel $\kappa(x_i, x_j)$, we obtain:

$$k(x_i, x_j) = \exp\{-\frac{1}{2\sigma^2}(\kappa(x_i, x_i) + \kappa(x_j, x_j) - 2\kappa(x_i, x_j))\} \tag{8}$$

Sigmoid kernel is another common kernel function as follows:

$$k(x_i, x_j) = \tanh(a * x_i^T x_j + b) \tag{9}$$

whose Gram matrix in general is not positive semi-definite. This form of kernel has, however, been used in practice (Vapnik, 1995), possibly because it gives kernel expansions such as the support vector machine a superficial resemblance to neural network models. For $a > 0$, we can view as a scaling parameter of the input data, and b as a shifting parameter that controls the threshold of mapping. For $a < 0$, the dot-product of the input data is not only scaled but reversed.(Ref.1). The value of $a$ is usually equal to reverse of number of features.

## RELATED WORKS

Evans and Ishikawa [6] presented an algorithm to measure growth rate and swing rate of tip angle automatically and they implemented it by ADAPT[1] software. Miller [14] provided an accurate method to obtain mid line of plant image. Dashti [3] undermined the significance of phenotypic traits that are implicit in patterns of dynamics in plant root response to sudden changes of its environmental conditions. Siahpirani[15] designed an PCA-based algorithm to obtain midline of root and number of root hairs. Miller [13] employed LDA and wavelet differentiation to analyse root gravitropism phenotypic traits. D.Brook [4] differentiated plasticity of root plant in different conditions.

## ANALYSIS OF MIXTURE OF EXPERT AND NEGATIVE CORRELATION LEARNING MODELS

In this section, the ME and NCL methods are investigated and reviewed.

### A.ME

The ME method was introduced by Jacobs et al.[9,10]. The authors examined the use of different error functions in the learning process for expert networks in the ME method. Jacobs proposed making NN experts local in different distributions of data space; as a result, the increased diversity among the experts led to improvements in the performance of this method. Various error functions were investigated with respect to a performance criterion and then Jacobs introduced a new error function based on the negative log probability of generating the desired output vector under the mixture of Gaussian models:

$$E = -\log \sum_j g_j \exp(-\frac{1}{2}(y - O_j)^2), \tag{10}$$

where $g_j$ is the proportional contribution of expert $j$ to the combined output vector and $O_i$ and $y$ are the actual and desired outputs of the $i^{th}$ NN, respectively.

To evaluate this error function, its deviation with respect to $i^{th}$ expert is analysed:

$$\frac{\delta E}{\delta O_i} = -\left[\frac{g_i \exp(-\frac{1}{2}(y - O_i)^2)}{\sum_j g_j \exp(-\frac{1}{2}(y - O_j)^2)}\right](y - O_i). \tag{11}$$

---
[1] Automatic Degree And Position Technique

According to the derivation term, the training of each expert is based on its individual error. Moreover, the weight-updating factor for each expert is proportional to the ratio of its error value to the total error. These two features in the proposed error function that cause the localization of each expert in their corresponding and authors claimed that the ME method has better efficiency with this error function.

In addition, a gating network is used to complete a system of competing local experts. The learning rule for the gating network attempts to maximize the likelihood of the training set by assuming a Gaussian mixture model in which each expert is responsible for one component of the mixture.

The ME method has special characteristics that distinguish it from the other combining methods. This method differs from the others due to its dynamic combination method. In the literature on combining methods, ME refers to the methods in which complex problems based on a "divide and conquer" approach are partitioned into a set of simpler sub problems and are distributed among the experts. In this method, instead of assigning a set of fixed combinational weights to the experts, as described previously, an extra gating network is used to compute these weights from the inputs dynamically.

*B. NCL*

In neural network ensemble methods, the individual networks are usually trained independently. One of the disadvantages of such an approach is the loss of interactions among the individual networks during the learning process. It is thus possible that some of the independently designed individual networks contribute slightly to the whole ensemble.

Liu and Yao [11] proposed the NCL method that trains NNs in the ensemble simultaneously and interactively through the correlation penalty terms in their error functions. In NCL, the error function of the $i^{th}$ NN is expressed by the equation:

$$E_i = \frac{1}{2}(y - O_i)^2 + \lambda P_i, \qquad (12)$$

where $O_i$ and $y$ are the actual and desired outputs of the $i^{th}$ NN, respectively. The first term in (3) is the empirical risk function of the $i^{th}$ NN. The second term, $P_i$, is the correlation penalty function, which can be expressed as:

$$P_i = (O_i - O_{ens}) \sum_{i \neq j} (O_j - O_{ens}), \quad (13)$$

where $O_{ens}$ is the average of NNs outputs in the ensemble. Here, $P_i$ can be regarded as a regularization term that is incorporated into the error function of each ensemble network. This regularization parameter provides a convenient way to balance the bias-variance-covariance trade-off [11]. This term is meant to quantify the amount of error correlation, so it can be minimized explicitly during training, which leads to negatively correlated NNs. The term $\lambda$ is a scaling coefficient parameter that controls the trade-off between the objective and penalty functions. When $\lambda = 0$, the penalty function is removed, resulting in an ensemble in which each network trains independently of the others, using simple Back-Propagation (BP). Therefore, the interaction and correlation among NNs of the ensemble is controlled explicitly by the value of $\lambda$. The minimization of this penalty function encourages different individual networks in an ensemble to learn different parts or aspects of the training data so that the ensemble can better learn the whole training dataset.

*C. Comparison the strengths and weaknesses*

In this part we compare the strengths and weaknesses of ME and NCL models.

Frist, we regard common characteristics of these two models. Both ensemble methods train base NNs in a process with multiple communication and cooperation among experts simultaneously. As already mentioned, different and special functions of these models have characteristics that lead experts to learn sub problems or different aspect of problem in a comparative and cooperation process, and finally the hybrid system learns all the data space. In other word, the error functions of these methods have characteristics that base NNs are produced with bias that possess negative correlation by dividing data space between experts explicitly [8].

Beside these similarities, there are some differences. Hybrid system contains two major parts. In fist part, training of base NNs, NCL has better efficiency than ME. Regularization term, which is used in this method, provides a convenient way to balance the bias-variance–covariance trade-off and thus improves the generalization ability, whereas ME does not include such control over the trade-off. On the other hand, one of the superiors of base MEs is its unique technique to combine output of experts.ME uses a trainable combiner that, according to the input *x*, dynamically selects the best expert(s) and combines their outputs to create the final output. The combining function of ME includes a dynamic weighted average in which the local competences of the experts with respect to the input are estimated by the weights produced by the gating network.

### GATING NETWORK TO COMBINE NCL EXPERTS OUTPUTS

According to complement characteristics of these two models, an integrated system can reserve strengths points and reduce their weaknesses. Based on this idea, the proposed hybrid system [5] contains two stages of training. In first stage, NNs are trained by NCL algorithm and in second stage; gating network is used for combining base NNs which is resulted from pervious stage. From another point of view, we can regard this method as improved NCL method. In NCL, error function orients base NNs to train different aspects or parts of problem. The local expertise characteristic of base NNs of NCL should be regarded in their combination method. Using gating network, as a combiner of base NNs output, can provide this required ensemble NN characteristic in NCL method. In this method, after training of neural networks by NCL, in the next stage, gating network is trained to model local expertise and to combine them; therefore, this method is called "using gating network to combine NCL experts".

For implementation of this idea, gating network should be train by some target points that show the efficiency and local expertise of experts in different sub problems. Like base mixture of expert $h_{G-NCL}$ is proposed to measure local and proportional expertise of base NNs.

$$h_{G-NCl,i} = \frac{\exp(-\frac{1}{2}(y-O_i)^2)}{\sum_j \exp(-\frac{1}{2}(y-O_j)^2)}. \quad (14)$$

If these values be in used as the target points to train gating network, gating network can estimate base NNs local expertise proportion in accordance to system input. With respect to $h_{G-NCL}$ and calculating target points in (5), error function of gating network is expressed as follows:

$$E_{G-NCL} = \frac{1}{2}(h_{G-NCL} - g)^2. \quad (15)$$

This error function is used to train Multilayer Perceptron network with Back Propagation algorithm. After output production of gating network, $O_{g,i}$, output of neuron $i^{th}$, softmax function applies to calculate $g_i$:

$$g_i = \frac{\exp(O_{g,i})}{\sum_{j=1}^{N} \exp(O_{g,j})}. \quad (16)$$

Here, the $g_i$ values are nonnegative and sum to unity, and they can be interpreted as estimates of the prior probability that expert *i* can generate the desired output *y*. Learning algorithm for updating weights of input layer to hidden layer and hidden layer to output layer based on Back Propagation is as follow:

$$\nabla w_{yg} = \eta_g (h_{G-NCL} - g)(O_g(1-O_g))O_{hg}^T \qquad (17)$$

$$\nabla w_{hg} = \eta_g w_{yg}^T (h_{G-NCL} - g)(O_g(1-O_g)) \\ * O_{hg}(1-O_{hg})x_i \qquad (18)$$

where $\eta_g$ is learning rate, $g$ gating network output after applying softmax function, $w_{hg}$ and $w_{yg}$ are weights of hidden layer and weights of output layer, recpectively. $O_{hg}^T$ is transposed vector of $O_{hg}$ neurons outputs of hidden layer of gating network. Finally, to combine of NCL experts outputs, the gate assigns a weight $g_i$ as a function of $x$ to each of the experts' output, $O_j$ and the final combined output of the ensemble is:

$$O_T = \sum_{j=1}^{N} O_i g_i. \qquad (19)$$

The idea of using gating network is an efficient method to combine output of experts' based on local expertise. In this idea, weights which are assigned to each expert output are estimated based on local expertise rate. In this approach, the combining weights are estimated dynamically from the inputs based on the different competences of each expert regarding different parts of the problem. Hence, the combination of NCL experts using this approach is superior to the previous static methods. The whole process is show in Fig.1a, b. The first stage is shown in Fig. 1a. In the first training stage, the expert networks are trained using the NCL error function. Fig. 1b shows the second stage of the training algorithm. After training the NCL experts, in the second stage, a gating network is trained to model the local competence of the NCL expert.

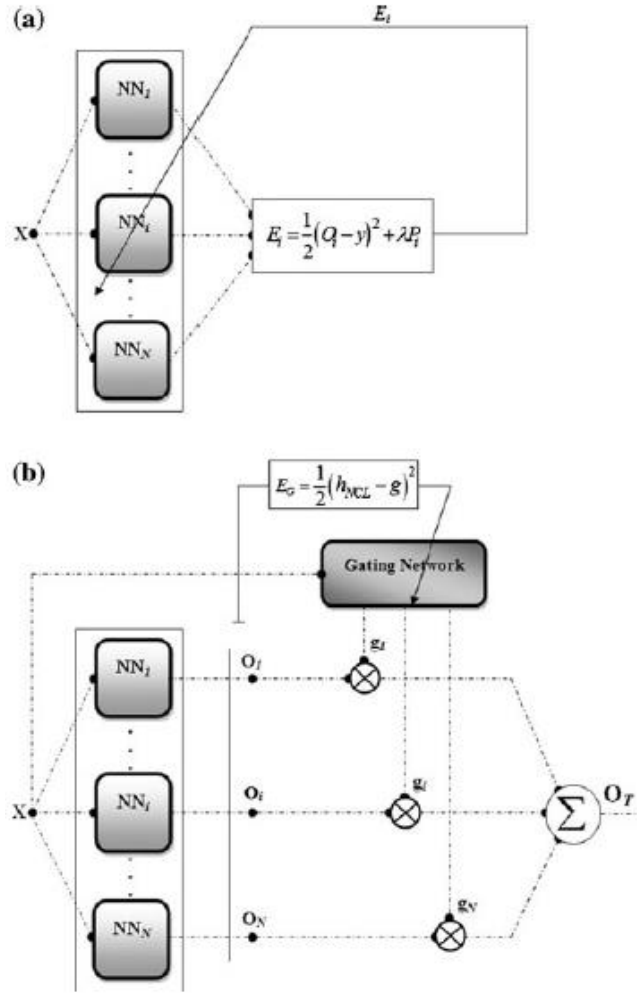

Fig. 1 a, b. Diagram of the two training stages of Gated NCL.

## INCORPORATION OF THE NCL TRAINING ALGORITHM INTO ME

ME and NCL methods use different error functions to incorporate negative correlation between the experts.
Although ME can produce negatively correlated experts, it does not include a control parameter like NCL to adjust this parameter explicitly and so make a near-optimal balance in the bias-variance-covariance trade-off. To introduce this advantage of NCL into the training algorithm of ME, we incorporated this control parameter in the ME error function. The modified error function of ME was obtained by adding the penalty term from NCL to the error function. The proposed method is named Mixture of Negatively Correlated Experts (MNCE). The new error function thus takes the form of Equations (20) and (21):

$$E = -\log \sum_j g_j \exp(-\frac{1}{2}(y - O_j)^2 + \lambda P_j), \qquad (20)$$

$$P_i = (O_i - O_{ens}) \sum_{i \neq j} (O_j - O_{ens}), \qquad (21)$$

The introduced penalty term, similarly to NCL, provides a control parameter leading to a near-optimal balance in the bias-variance-covariance trade-off. In this ensemble architecture, each expert network is an MLP network with one hidden layer that computes an output $O_i$ as a function of the input vector $x$ and the weights of hidden and output layers with a sigmoid activation function.

To train MLP experts based on the new error function using the BP training algorithm, the weights for each expert $i$ are updated according to the following rules:

$$h_{MNCE,i} = -\left[ \frac{g_i \exp(-\frac{1}{2}(y - O_i)^2 + \lambda P_i)}{\sum_j g_i \exp(-\frac{1}{2}(y - O_j)^2 + \lambda P_i)} \right]. \qquad (22)$$

$$\nabla w_{y,i} = \eta_e h_{MNCE,i} \left[ (y - O_i) - \lambda \frac{\partial P_i}{\partial O_i} \right] (O_i(1 - O_i)) O_{hi}^T \qquad (23)$$

$$\nabla w_{h,i} = \eta_e h_{MNCE,i} w_y^T \left[ (y - O_i) - \lambda \frac{\partial P_i}{\partial O_i} \right] (O_i(1 - O_i)) O_{hi}(1 - O_{hi}) x_i \qquad (24)$$

$$\frac{\partial P_i}{\partial O_i} = \left[ g_i \sum_{j \neq i} (O_j - \overline{O}) + g_i (M - 1)(O_i - \overline{O}) \right] \qquad (25)$$

$$\overline{O}(n) = \frac{1}{M} \sum_{i=1}^{N} O_i \qquad (26)$$

where $\eta_e$ is the learning rate, $\lambda$ is the NCL control parameter, $g_i$ is the $i^{th}$ output of the gating network after applying the softmax function, $w_h$ and $w_y$ are the weights of the inputs to the hidden layers and those of the hidden to the output layers of the expert networks, respectively.

$O_h^T$ is the transpose of $O_k$, the outputs of the hidden layers of the expert networks. Similar to original ME, the gate is composed of two layers: the first layer is a MLP network, and the second layer is a softmax nonlinear operator.

Thus, the gating network computes $O_g$, which is the output of the MLP layer of the gating network, then applies the softmax function (Eq. 16).
According to the modified ME error function (Eq. 20), the modified error function of the gating network can be written as:

$$E_{G,MNCE} = \frac{1}{2}(h_{MNCE} - g)^2 \qquad (27)$$

Based on this error function, the weights of the gating network in the MNCE method are determined using the BP error algorithm according to the following rules:

$$\Delta w_{yg} = \eta_g (h_{MNCE} - g)(O_g(1-O_g))O_{hg}^T \qquad (28)$$

$$\Delta w_{hg} = \eta_g w_{yg}^T (h_{MNCE} - g)(O_g(1-O_g))O_{hg}(1-O_{hg})x_i \qquad (29)$$

Finally, to combine the experts' outputs, the gate assigns a weight $g_i$ as function of $x$ to each of expert's output $O_j$, and the final mixed output of the ensemble is:

$$O_T = \sum_j O_j g_j \qquad (30)$$

The MNCE is composed of the experts and a gating network. The expert training process and the gating network work simultaneously to minimize the modified error functions. The experts compete to learn the training patterns, and the gating network mediates the competition. The added control parameter $\lambda P_i$ provides an explicit control for efficiently adjusting the measure of negative correlation between the experts. The structure of MNCE and its simultaneous training algorithm for the experts and gating network are shown in Figure 2.

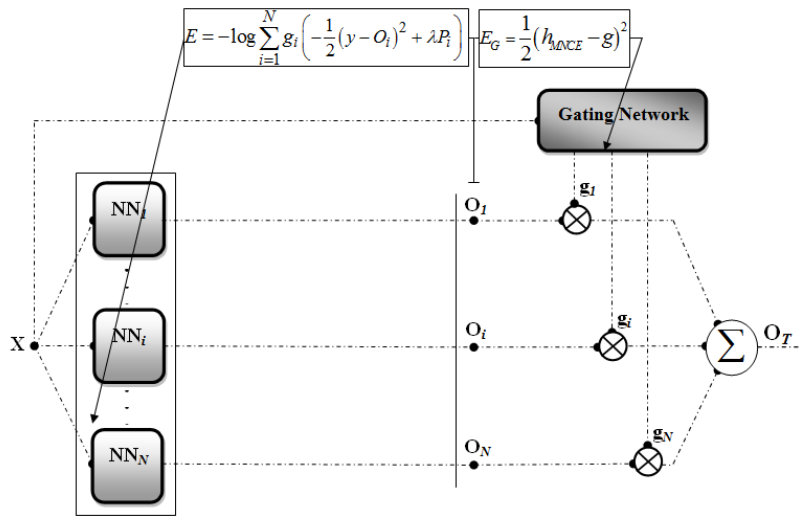

**Fig. 2.** Diagram of the Mixture of Negatively Correlated Experts (MNCE).

According to this MNCE training procedure, in the network's learning process, the expert networks compete for each input pattern and the gating network rewards the winner of each competition with stronger error feedback signals. This competitive learning procedure leads to a localization of the experts into possibly overlapping regions. Additionally, incorporating the control parameter of NCL into the error function of ME provides an explicit control for efficiently adjusting the measure of negative correlation between the experts. Thus, it yields a better balance in the bias-variance-covariance trade-off, which improves generalization ability.

## USING VELOCITY AND ACCELERATION FOR FEATURE EXTRACTION

In this part we explain our feature extraction approach which is applied on time series dataset. Frist we use Principal Component Analysis (PCA) to reduce the dimensions of data in dataset. We also obtain principal components of data in this stage. In datasets that have some motions we can calculate velocity and acceleration of motions in consecutive data as follows:

$$Velocity(i) = P(i)(j-1) + P(i)(j), \quad (11)$$

$$Acceleration = Velocity(i)(j+1) - Velocity(i)(j), \quad (12)$$

$$for\ i = 1,...,m\ and\ j = 1,...,n-1$$

where $P$ is the main matrix, $m$ is the number of samples and $n$ is the length of each samples. We used sliding window, a method used for data mining in time series in machine learning, to improve classification rate. Sliding window moves throughout the feature vector and the classifiers in these slices will be trained. In this way we can find best slices to obtain best features to classify dataset.

## EXPERIMENTAL

To evaluate the results, first we define our dataset. Then, the approach for dimension reduction is principal component analysis. In next step, we define our feature vector and give it to different classifiers to classify our dataset.

### A. Dataset

Main purpose of studying Arabidopsis Thaliana model plant is to find the operation of each 25000 genes. An extensive and indexed library of mutated T-DNA input leads to use reverse-genetic to find these operations. When a mutation causes observable change in phenotype to obtain clue of disabled gene function, reverse-genetic can be efficient. However, many disabled gene in Arabidopsis Thaliana do not make any observable changes in phenotype.

### B. Size of seeds

Seeds of Arabidopsis Thaliana were screened based on size $255\mu m^2$, $280\mu m^2$, $300\mu m^2$, $355\ \mu m^2$. Seeds which are between $255\mu m^2$ - $280\mu m^2$, are in small group, and seeds which are $300\mu m^2$ - $355\ \mu m^2$ are in large group. Seeds grain vertically in special circumstances between 2 and 4 days.

### C. Genetic Mutation

Seeds of plant which obtain mutated T-DNA in GLR.3.3 gene are provided by Salk association. Used product are Salk_040458 (glr3.3-1, mutated second axon) and Salk_066009 (glr3.3-2, mutated first antron) and seeds are divided by the method identified in pervious section.

### D. Genetic Mutation

As already discussed, dataset contains groups in Table 1.It is necessary for classification that at least one wild type species places against mutated type species. For this purpose, groups place against each other what is demonstrated Table 2.

Each group contains some samples which each sample contains 300 pictures. Resolution of each frame is 700*900 pixels that can be shown by a 630000-ordered pair which means each frame belongs to a 630000 dimensions. So the dimension of each sample is─630000 ×300 = 189000000 this dimension is so large to direct classification. We use principle component analysis for dimension reduction. As already mentioned, the dimensions of these points are 630000 that we transform them to the 30-40 dimensions. It can be argued that the difference between two consecutive frames is very small. It means the seed tip moves just few pixels, so this difference is small in low dimension too. As a result, these determining points of each frame emerges a curve in space. This curve is a dynamic sequence of root growth. However, based on experiments used the curve with this length, we cannot obtain helpful answers. We use 2D array to show this curve that multiple pair ordered of each point place in one direction and time places in oppose direction.

TABLE I
DATA PROFILE USED AS DATASET

| Sample Number | Seedling age(day) | Mutated gene | Seed size | Group |
|---|---|---|---|---|
| 20 | 2 | GLR 3.3.1 | Large | 331L2 |
| 18 | 2 | GLR 3.3.2 | Large | 332L2 |
| 18 | 2 | GLR 3.3.1 | Small | 331S2 |
| 18 | 2 | GLR 3.3.2 | Small | 332S2 |
| 14 | 4 | GLR 3.3.1 | Large | 331L4 |
| 16 | 4 | GLR 3.3.2 | Large | 332L4 |
| 16 | 4 | GLR 3.3.1 | Small | 331S4 |
| 15 | 4 | GLR 3.3.2 | Small | 332S4 |
| 27 | 2 | Wild type | Large | wtL2 |
| 23 | 2 | Wild type | Small | wtS2 |
| 22 | 4 | Wild type | Large | wtL4 |
| 23 | 4 | Wild type | Small | wtS4 |

TABLE II
GROUPS FOR CLASSIFICATION

| Wild Type Group | Mutated Type Group |
|---|---|
| wtS2 | 331S2 |
| wtS2 | 332S2 |
| wtS4 | 331S4 |
| wtS4 | 332S4 |
| wtL2 | 331L2 |
| wtL2 | 332L2 |
| wtL4 | 331L4 |
| wtL4 | 332L4 |

For each pair that contains a mutated group and a wild type group 30 through 40 principal components is considered. Then a 2D array which is corresponding of merging of number of frame and number of principal component is obtained by a 3D matrix of "number of samples" × "number of frames" × "number of principal component". Velocity and acceleration is obtained from this matrix and is added to end of this matrix.

For classification rate of SVM we used cross-validation method with 5 folds. MLP neural networks are used with one hidden layer with 4 neurons in it when gating network NCL experts' method is applied. All methods are trained by Back Propagation algorithm. In Gated-NCL learning rate for experts and gating networks are 0.15 and 0.1, respectively. $\lambda$ is obtained by trial and error procedure in Gated-NCL and MNCE. Also, here we used cross-validation method with 5 folds for correct rate of this classifier. On average the best results were obtained using a window length of 40 that it is used for all other experiments. Results are shown in Table 3.

Feature Extraction and Classification Procedure for Analysis of Root Growth

1. Compute Feature Vector with PCA method
2. Make 2D array 'P' by merging of Number of Frame and Number of PCs
    - Compute velocity and acceleration   growth rate of each frame of samples
    - add these features to the end of the 'P'
3. Move Sidling window on feature vector end of file
     Do training and testing operation

4. output : lowest  error rate and best frames for training
   of classifiers

## CONCLUSION

In this paper, we proposed a new approach for feature extraction base on velocity and acceleration of motion and employed a method based on NCL and ME applicable for classification.

We used a Gated-NCL and MNCE hybrid system by regarding weaknesses and strengths of NCL and ME models. NCL encourages experts by "divide and conquer" principal that consider different parts and aspects of data space.

We also employed different kinds of SVM classifiers to compare their efficiency with each and other ensemble methods. According to result, the MNCE hybrid system can obtain better performance than GNCL. Moreover, Gaussian kernel worsens the efficiency of SVM. Finally, sigmoidal kernel could obtain best classification error rate between our classifiers. In addition, these hybrid system spend more time than kernel SVMs, because dividing the problem to some sub problems and then integrating the result is time-consuming.

In future work, we want to apply semi-supervised algorithms. SVM will be exploited with generative models such as Parzen windows and probabilistic neural network (PNN) in order to improve the training process and reduce the time of testing. It is also interesting to study for using different kernel functions for the SVM with semi-supervised algorithms.

TABLE III
COMPARISON OF CLASSIFIERS ERROR RATE

| Sigmoid – SVM | Gaussian-SVM | Linear-SVM | MNCE | ME | Gated-NCL | NCL | Best Frames | Wild Type | Mutated Type |
|---|---|---|---|---|---|---|---|---|---|
| %12.5 | %25.23 | %19.52 | %16.66 | %19.34 | %18.27 | %20.48 | 91-131 | wtS2 | 331S2 |
| %11.82 | %17.44 | %12.40 | %11.92 | %12.02 | %12.14 | %12.21 | 249-289 | wtS2 | 332S2 |
| %6.48 | %24.21 | %17.11 | %7.89 | %9.44 | %9.02 | %10.15 | 2-42 | wtS4 | 331S4 |
| %15.04 | %27.09 | %22.16 | %17.69 | %21.11 | %19.67 | %20.78 | 255-295 | wtS4 | 332S4 |
| %19.15 | %26.66 | %19.97 | %19.80 | %20.40 | %19.91 | %20.25 | 253-293 | wtL2 | 332L2 |
| %5.44 | %28.11 | %24.20 | %6.82 | %7.19 | %8.87 | %9.80 | 257-297 | wtL2 | 332L2 |
| %19.54 | %33.14 | %29.23 | %23.13 | %27.18 | %26.12 | %29.44 | 251-291 | wtL4 | 331L4 |
| %19.43 | %35.57 | %31.75 | %20.21 | %27.43 | %22.87 | %26.32 | 250-290 | wtL4 | 332L4 |